# An innovative method to increase the efficiency of AI-based fraud detection model in automobile insurance using TNF based spectral embedding

**Rohan Yashraj Gupta*, Lalith Srikanth Chintalapati, Satya Sai Mudigonda, Pallav Kumar Baruah, Raghunatha Sarma Rachakonda**

Department of Mathematics and Computer Science,
Sri Sathya Sai Institute of Higher Learning,
Vidyagiri, Puttaparthi, Andhra Pradesh, India.
**Corresponding author*: rohanyashrajgupta@sssihl.edu.in

**Abstract**

Fraud detection is an important area of research in the insurance business due to its financial implications. The primary aim of a fraud detection model is to identify fraud and non-fraud cases with high accuracy along with other important metrics such as Sensitivity, Specificity, Precision, F1-score, False Positive Rate, False Discovery Rate, AUC etc. To achieve this, we need to explore a suitable classification model to identify fraud and non-fraud cases. In this work, we have used auto insurance data set and explored classification models such as Decision Tree (DT), Random Forest (RF), XGBoost, LightGBM and Gradient Boosting Machine (GBM). To overcome the problem of data imbalance, we have employed MWMOTE and TGAN techniques. We have used Topological Node Feature(TNF) based spectral embedding for low dimensional data representation along with some popular embedding methods like MDS, Isomaps and t-SNE. After studying all the 65 possible combinations of these models, we have proposed an innovative method for effective automobile insurance fraud detection. For the given dataset, our results show that using a combination of MWMOTE as a data imbalance handling technique (Phase I), TNFSE2 as data embedding (Phase II) and Random Forest as classification (Phase III) provides the best result in comparison to all other combinations. This work also highlights the efficacy of TNF based spectral embedding in automobile insurance dataset

## 1. Introduction

Insurance companies lose a large amount of their revenue due to insurance fraud. There are various measures that they take to ensure that such losses are minimized. According to the Federal Bureau of Investigation, the total of $40 billion per year is lost in non-health insurance fraud and about 4% of the total money made by insurance is lost towards insurance fraud (“Insurance Fraud — FBI,” n.d.). According to the National Insurance Crime Bureau (NICB), there has been a continuous rise in questionable claims year on year, with a 34 percent increase between 2008 and 2011. Approximately, 21% - 36% of auto-insurance claims are suspicious in nature, but only less than 3% of it is prosecuted (Nian, Zhang, Tayal, Coleman, & Li, 2016).

Traditional fraud detection models which are heavily dependent on auditing and expert inspection is no longer a viable option. These models are costly, inefficient, time-consuming and require lots of human intervention. In recent time, various organizations are adopting data mining and machine learning techniques in analyzing a large amount of data and building fraud detection models.

Efforts are being made by various researchers in this domain to build a better fraud detection model. Viaene et.al. used Bayesian learning neural networks for auto claims fraud detection (VIAENE, DEDENE, & DERRIG, 2005). Bermudez et. al. introduced an asymmetric Bayesian dichotomous logit model for finding the fraudulent insurance claims found in a Spanish insurance market (Bermúdez, Pérez, Ayuso, Gómez, & Vázquez, 2008). Subelj et. al. used a graph-based social network model to identify frauds in automobile insurance (Šubelj, Furlan, & Bajec, 2011). They have developed an Iterative Assessment Algorithm (IAA) that was based on Graph Components for identifying suspicious claims. Xu et. al. used a random rough subspace-based neural network ensemble for insurance fraud detection (W. Xu, Wang, Zhang, & Yang, 2011). Sunderkumar et. al. used a One-Class Support Vector Machine (OCSVM) as an under-sampling technique to handle the class imbalance problem (Sundarkumar, Ravi, & Siddeshwar, 2015). Nian et. al. provided an unsupervised model for auto insurance fraud detection (Nian et al., 2016). Subudhi et. al. proposed the use of fuzzy c-means clustering for balancing the dataset and used a threshold technique to identify whether the majority of samples are outliers or not (Subudhi & Panigrahi, 2020).

One challenge that is faced when training such model is the number of features in the dataset. Some widely used models to reduce the dimension of the dataset include Multidimensional scaling (MDS) which is used to represent the level of similarity or dissimilarity present in the data as distances in a geometric space (J. Kruskal & Wish, 2011). An extension of this was proposed by Wang et. al., Isomaps (Wang & Wang, 2012). In this model, the distance is calculated as geodesic distances which are defined as the number of edges in the shortest path connecting the two points in the graph. t-Distributed Stochastic Neighbor Embedding (t-SNE) converts the affinities of the data points into probabilities. The affinities in the embedded space are represented by t-distribution (Van Der Maaten & Hinton, 2008). Belkin et. al. proposed the Laplacian Eigenmap (LE) model to find an appropriate representation of data (Belkin & Niyogi, 2002).

Traditionally in the field of Spectral Clustering, the pairwise affinity is defined using Gaussian kernel weighted Euclidean distance. Zhang et. al. proposed an affinity measure based on modified Gaussian similarity using Common Nearest Neighbors (CNN) (Zhang, Li, & Yu, 2011). Arias et. al. proposed a spectral clustering algorithm based on features from local PCA (Arias-Castro, Lerman, & Zhang, 2017). To capture the local properties of the data effectively, Chintalapati et. al. have utilized Topological Node Features (TNFs) from the graph representation of the data and proposed a novel spectral clustering technique (Chintalapati & Rachakonda, 2019). They have used the work of Dahm et. al. for building the affinity measure that effectively captures local information (Dahm, Bunke, Caelli, & Gao, 2015). We have used the TNFs and proposed a novel embedding method which is referred to as TNFSE1 and TNFSE2 in this work.

In this work, we have also used a three-phased method of fraud detection which consists of handling data imbalance in phase I, data embedding in phase II and classification models in phase III. We have used a various combination of models in each of these phases and proposed the most effective method for fraud detection in automobile insurance.

The rest of the paper is organized as follows: Section 2 presents the methodology. Section 3 details the data description. Section 4 explains the performance metrics which are used to evaluate the performance of the models. Section 5 discusses the results of various models. Finally, Section 6 presents the conclusion of the work and ideas for future work.

## 2. Methodology

In this section, we will discuss a three-phased method for building a fraud detection model. The schematic representation of our approach is shown in Figure 1. In phase I, we have handled the data imbalance using two over-sampling techniques - MWMOTE and TGANs as shown in the works of Gupta et. al. and Rai. et. al.(Gupta, Mudigonda, & Baruah, 2021; Rai, Baruah, Mudigonda, & Kandala, 2018) (Ashrapov, 2020; L. Xu & Veeramachaneni, 2018) (Gupta et al., 2021). Here, the minority samples are synthetically generated to balance the two classes in the dataset. In phase

II, we have performed data embedding using a novel TNF based Spectral Embedding (TNFSE1 and TNFSE2) and also performed a comparative study with respect to other popular embedding methods like Isomap, MDS and t-SNE. In phase III, we have used various classification models such as Decision Tree (DT), Random Forest (RF), XGBoost, LightGBM and Gradient Boosting Machine (GBM).

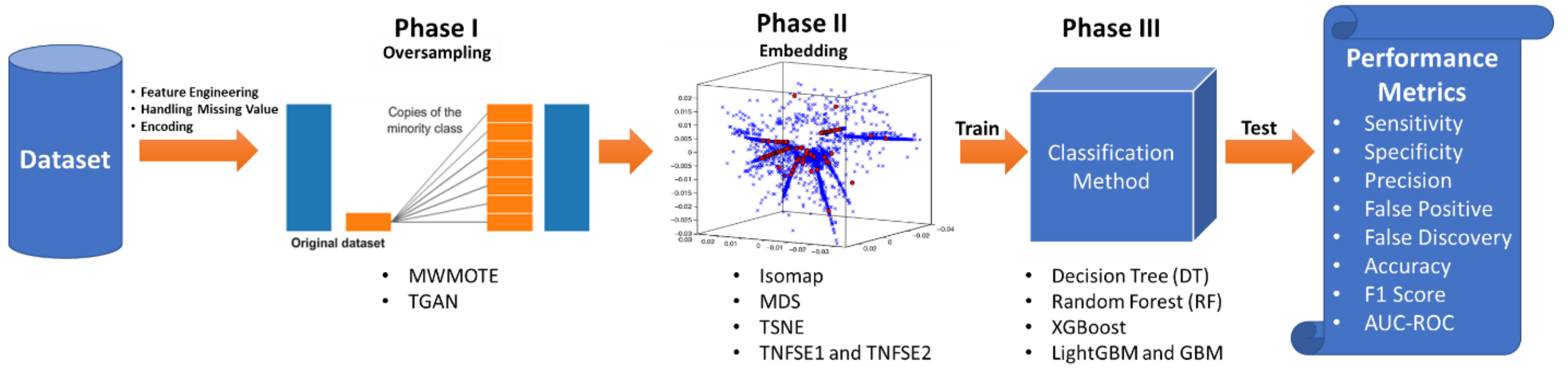

**Figure 1 - Proposed method**

## 2.1 Phase I

In this phase, we have used Majority Weighted Minority Oversampling Technique (MWMOTE) and Tabular Generative Adversarial Networks (TGANs) to oversample fraudulent claims in the dataset. In MWMOTE, weights are assigned to the minority classes such that the oversampled claims are inside some minority cluster (Barua, Islam, Yao, & Murase, 2014). TGANs is a neural network based generative model which is trained to learn the distribution of the minority samples (Ashrapov, 2020; Gupta et al., 2021; L. Xu & Veeramachaneni, 2018). The model is said to be trained when the discriminator is no longer able to distinguish between real and synthetic dataset. The trained model is then used to generate minority samples.

## 2.2 Phase II

In this phase, embedding models are used for low-dimensional representation of the dataset. It makes the similarities between data points more apparent, and the classification becomes more accurate. We have performed data embedding using a novel TNF based Spectral Embedding (TNFSE1 and TNFSE2) and also performed a comparative study with respect to other popular embedding methods like Isomap, MDS and t-SNE.

A novel embedding method using Topological Node Features (TNF) is used in this work. The following are the TNFs proposed by Chintalapati et.al. (Chintalapati & Rachakonda, 2019; Dahm et al., 2015):

1. The degree of a node $p$ in G is given by $d(p) = |\{q \in V \mid (p,q) \in E\}|$, where the vicinity graph G = (V, E). V is the set of vertices and E is the set of edges connecting two points if the Euclidean distance between them is less than a pre-assigned parameter $\varepsilon = 0.5$
2. Clustering Coefficient ($\phi_p$) of a node $p$ is the number of edges among the vertices in $N(p)$, i.e. $\phi_p = |\{(p,q) \in E \mid p,q \in N(p)\}|$, where the first neighborhood $N(p)$ represents the set of all the nodes which are connected to a node $p$. It is defined as $N(p) = \{q \in V \mid (p,q) \in E\}$
3. The Summation Index ($SI$): This is a measure of the structural information at a point. $SI$

summarizes the degree TNF of neighborhood points into one value. $SI_i$ is recursively defined in the work of Dahm et al., 2015 as:

$$SI_i(p) = degree(p) \qquad if\ i = 0 \qquad (3)$$
$$= \sum_{q \in N(p)} SI_{i-1}(q) \qquad if\ i > 0$$

$V_p$ for every node $p$ is defined as $V_p = (SI_0, SI_1, SI_2)$

Using these TNFs, the affinity matrix is calculated as shown in algorithm 1

**Algorithm 1: The TNF based affinity algorithm**

---

**Input:** Given a set of points $S = p_1, \ldots, p_n$ in $R^l$, sigma value
**Output:** Affinity matrix A

1. model the data points as a graph G
2. At each node p, calculate the following TNFs:
   (a) Degree of node ($d_p$)
   (b) Clustering coefficient ($\phi_p$)
   (c) SI vector $V_x$ = ($SI_0$, $SI_1$, $SI_2$)

3. We define the similarity $\alpha_{ij}$ between any two nodes $p_i, p_j$ as:

$$\alpha_{ij} = exp(\frac{-||\ p_i - p_j\ ||^2 \delta_{ij}}{2\sigma^2}). \eta_{ij}. (1 + \frac{1}{1 + log(1 + \zeta_{ij})})$$

where $\delta_{ij} = abs(\phi_i - \phi_j)$, $\eta_{ij}$ is the number of common points between $N(p_i), N(p_j)$, $\sigma$ is the scale parameter of the Gaussian function and $\zeta_{ij}$ is the Euclidean distance between local structural information of $p_i$ and $p_j$. The affinity matrix gives the idea about how closely are the claims related to each other. Thus, $\alpha_{ij} > \alpha_{ik}$ indicates that the claims i and j are more closely related than the claims i and k. After the affinity matrix is created spectral embedding of data is done (Belkin & Niyogi, 2002). Algorithm 2 details how the Eigen spectrum of the Laplacian matrix is used to find the spectral embedding of the data.

**Algorithm 2: Spectral Embedding algorithm**

---

**INPUT:** Given a set of points $S = p_1, \ldots, p_n$ in $R^l$, parameters $k$, $\sigma$.
**OUTPUT:** $k$ clusters of the given data

1. The affinity matrix $A \in R^{n \times n}$ is constructed using such that affinity is given as $A_{ij} = exp(-||p_i - p_j||/2\sigma^2)$
2. Degree matrix D is defined as the diagonal matrix whose $(i, i)$- element is the sum of A's i-th row. Construct Laplacian matrix L as: $L = D^{-1/2} A D^{-1/2}$.
3. The k largest eigenvectors of L, $ev_1, ev_2, \ldots, ev_k$ (chosen to be orthogonal to each other in the case of repeated eigenvalues) are calculated, and the matrix $Y = [ev_1 \quad ev_2 \quad \ldots \quad ev_k] \in R^{n \times k}$ is constructed by stacking the eigenvectors as columns.
4. Normalize each of Y's rows to have unit length (i.e. $Y_{ij} = Y_{ij}/(\sum_j Y_{ij}^2\ )^{1/2}$) to obtain

spectral embedding of the data points in k dimension.

---

Using the Topological Node Features (TNF) based similarity proposed by Chintalapati and Rachakonda., (2019) (Chintalapati & Rachakonda, 2019) as $A_{ij}$ in Algorithm 1, a spectral embedding of the data points is obtained. This is referred to as TNFSE in this work. Here, there are two variations of TNFSE – using degree and clustering coefficient only is referred to TNFSE1 and using degree, clustering coefficient along with summation index is referred to as TNFSE2.

The proposed embedding methods TNFSE1 and TNFSE2 are tested against other popular embedding methods Multi-dimensional Scaling (MDS), Isomaps and t-SNE. Multi-dimensional Scaling (MDS) model is used to represent the level of similarity or dissimilarity present in the data as distances in a geometric space. (J. B. Kruskal, 1964b, 1964a; O'Connell, Borg, & Groenen, 1999). The distance between the pair of data points is measured as Euclidean distances. Isomap can be seen as an extension of MDS (Tenenbaum, 2000). In this model, the distance is calculated as geodesic distances which is defined as the number of edges in the shortest path connecting the two points in the graph. t-distributed Stochastic Neighbour Embedding (t-SNE) converts the affinities of the data points into probabilities. The affinities in the embedded space are represented by t-distribution. This is useful in revealing various attributes of the dataset like, structure, data that lying in multiple clusters and reduce the crowding the data points in the embedded space.

### 2.3. Phase III

The classification of claims into fraud and non-fraud is done using five machine learning models. In simple terms, Decision Tree (DT) learns from the data and fits an if-then-else decision rule into the dataset. The rules are then used for the classification of claims in fraud or non-fraud(Japkowicz & Stephen, 2002). Random forest is an ensemble model which is constructed using multiple decision trees (Tin Kam Ho, 1995; Žižka, Dařena, & Svoboda, 2019). This model is similar to decision trees with improved performance because of its ability to learn complex rules in the dataset. Gradient boosting is another machine learning model which is based on the ensemble of weak prediction models. There are various implementations of this in the literature (Guelman, 2012; Gupta, Sai Mudigonda, Kandala, & Baruah, 2019). Variation of GBM called LightGBM and XGBoost is used in this work along with GBM. LightGBM is gradient boosting framework which is fast and based on decision trees. XGBoost is different from GBM in the use of loss function.

The combination of all the models explained as part of Phase I, II and III are implemented and the results along with the business interpretation of the best performing model are shown in section 5.

## 3. Data Description & preparation

An automobile claims dataset, "carclaims.txt", was used in this work (Phua, Alahakoon, & Lee, 2004). The dataset has 15,420 claim records with 6% fraudulent claims. It has a total of 32 features, with 6 ordinal and 25 categorical features. The dataset was pre-processed to use it for training the model. Some features were engineered from the existing dataset for e.g. "accident date", "reporting date" and "reporting delay". There was one record with missing values, this had been discarded as it didn't have any significant impact on the performance of the model. The categorical features were encoded using one-hot encoding. The dataset used was divided into train and test in the ratio of 70:30 to test the performance of the model. The split was done randomly to ensure that the ratio of fraud and non-fraud cases in the train and test dataset coincides.

## 4. Performance Metrics

The model was evaluated using eight different metrics: Sensitivity, Specificity, Precision, False Positive Rate, False Discovery Rate, Accuracy, F1 Score and AUC-ROC (Powers, 2007). These metrics are used to measure the effectiveness and usefulness of the model, which are calculated as follows:

$$Sensitivity = \frac{Fraud\ claims\ identified\ as\ fraud}{Total\ fraud\ claims\ (actual)}$$

$$Specificity = \frac{Non-fraud\ claims\ identified\ as\ non-fraud}{Total\ non-fraud\ claims\ (actual)}$$

$$Precision = \frac{Fraud\ claims\ identified\ as\ fraud}{Total\ claims\ identified\ as\ fraud\ by\ the\ model}$$

$$False\ Positive\ Rate = \frac{Non-fraud\ claims\ identified\ as\ fraud}{Total\ non-fraud\ claims\ (actual)}$$

$$False\ Discovery\ Rate = \frac{Non-fraud\ claims\ identified\ as\ fraud}{Total\ claims\ identified\ as\ fraud\ by\ the\ model}$$

$$Accuracy = \frac{Total\ correct\ predictions\ both\ fraud\ and\ non-fraud}{Total\ claims}$$

$$F1\ Score = \frac{2 \times Precision \times Recall}{Precision + Recall}$$

$$AUC = Area\ under\ the\ ROC\ curve$$

## 5. Results and discussion

Using the methodology described in section 2 we performed three studies:

- **Study 1** - uses classification model (Decision Tree (DT), Random Forest (RF), XGBoost, LightGBM and Gradient Boosting Machine (GBM)) without applying data imbalance handling and embedding techniques (Table 1 - M1 to M5)
- **Study 2** - uses classification model mentioned in study 1 after applying data imbalance handling techniques (MWMOTE and TGANs) without using any embedding techniques (Table 2 - M6 to M15)
- **Study 3** - uses classification model after applying data imbalance as mentioned in study 2 and also apply embedding techniques (Isomaps, MDS, t-SNE, TNFSE1, TNFSE2) (Table 3&4 - M16 to M65)

### 5.1 Study 1 results

In study 1, the baseline (raw data) was used directly for building the classification model. The results obtained are shown in Table 1. It is observed that the overall performance of all the models (M1 to M5) is very poor. The sensitivity of all the models are very low (<0.2), this indicates that the model was not efficient in identifying the fraud cases. On the other hand, the specificity was very high (>0.9), which indicates that the model was good at classifying non-fraud claims. This is undesirable because the insurance companies would want sensitivity also to be higher along with specificity. It can also be observed that the false discovery rate is very high which means there is a high rate of false positive in the total number of claims predicted as fraudulent by the model. This would lead to an unnecessary investigation of the legitimate claims which is a very costly process. This observation can be attributed to the imbalanced nature of the data. This caused the model to be trained majorly on the non-fraudulent claims, which resulted in the model output to be skewed towards the non-fraudulent claims.

**Table 1 - Results of baseline dataset with various classifiers**

| Classifier | Model | AUC-ROC | Sensitivity | Specificity | Precision | False Positive Rate | False Discovery Rate | Accuracy | F1 Score |
|---|---|---|---|---|---|---|---|---|---|
| **DT** | **M1** | **0.5613** | **0.1829** | **0.9396** | 0.1711 | 0.0604 | 0.8289 | 0.8913 | **0.1768** |
| **RF** | **M2** | 0.5012 | 0.0081 | 0.9942 | 0.0870 | 0.0058 | 0.9130 | 0.9313 | 0.0149 |
| **XGBoost** | **M3** | 0.5100 | 0.0203 | 0.9997 | **0.8333** | **0.0003** | 0.1667 | 0.9372 | 0.0397 |
| **LightGBM** | **M4** | 0.5112 | 0.0244 | 0.9981 | 0.4615 | 0.0019 | 0.5385 | 0.9359 | 0.0463 |
| **GBM** | **M5** | 0.5121 | 0.0244 | 0.9997 | 0.8571 | **0.0003** | **0.1429** | **0.9375** | 0.0474 |

## 5.2 Study 2 results

In the study 2, data imbalance was handled using MWMOTE and TGANs. The balancing ensured that there were sufficient fraudulent claims for the model to be trained on. The results obtained are shown in Table 2. It can be observed that the AUC-ROC value of the models is now much higher in comparison to Table 1 for all the models. The results indicate that the data imbalance technique has significant improvement in the performance of the model. This is because, in this study, the fraudulent data were much higher compared to the previous study, thus the model is now good at classifying both the fraud and non-fraud claims. Of all the models, M6 to M15, the best performance of the model based on the AUC-ROC value is that of M7. The AUC-ROC value is observed to be 0.9900. This model is also good in terms of accuracy and F1-score with the values of 0.9901 and 0.9902 respectively. A high value of F1-score would mean that the model has a high value of sensitivity and precision. From the insurance companies' point of view, they would want a very less number of false-positives due to higher claims investigation cost. This is measured effectively by false positive rate and false discovery rate. M15 is better than any other models with respect to these two metrics.

**Table 2 - Results of MWMOTE & TGANs balanced dataset with various classifiers**

| Data Imbalance Technique | Classifier | Model | AUC-ROC | Sensitivity | Specificity | Precision | False Positive Rate | False Discovery Rate | Accuracy | F1 Score |
|---|---|---|---|---|---|---|---|---|---|---|
| **MWMOTE** | **DT** | **M6** | 0.9642 | **0.9997** | 0.9287 | 0.9341 | 0.0713 | 0.0659 | 0.9644 | 0.9658 |
| | **RF** | **M7** | **0.9900** | 0.9989 | 0.9811 | 0.9817 | 0.0189 | 0.0183 | **0.9901** | **0.9902** |
| | **XGBoost** | **M8** | 0.7962 | 0.9421 | 0.6502 | 0.7316 | 0.3498 | 0.2684 | 0.7970 | 0.8236 |
| | **LightGBM** | **M9** | 0.8751 | 0.9761 | 0.7740 | 0.8139 | 0.2260 | 0.1861 | 0.8757 | 0.8876 |
| | **GBM** | **M10** | 0.7969 | 0.9402 | 0.6535 | 0.7331 | 0.3465 | 0.2669 | 0.7977 | 0.8238 |
| **TGAN** | **DT** | **M11** | 0.9480 | 0.9503 | 0.9458 | 0.9436 | 0.0542 | 0.0564 | 0.9480 | 0.9469 |
| | **RF** | **M12** | 0.9664 | 0.9403 | 0.9924 | 0.9916 | 0.0076 | 0.0084 | 0.9670 | 0.9653 |
| | **XGBoost** | **M13** | 0.9705 | 0.9423 | 0.9986 | 0.9985 | 0.0014 | 0.0015 | 0.9711 | 0.9696 |
| | **LightGBM** | **M14** | 0.9701 | 0.9423 | 0.9978 | 0.9976 | 0.0022 | 0.0024 | 0.9707 | 0.9692 |
| | **GBM** | **M15** | 0.9705 | 0.9420 | **0.9989** | **0.9988** | **0.0011** | **0.0012** | 0.9711 | 0.9696 |

## 5.3 Study 3 results

To further improve the performance of the model, in study3, data embedding was done on the balanced dataset. The objective was to find the model which performs better than models M1 to M15. In this regard, we have performed data embedding using Isomap, MDS, t-SNE, TNFSE1 and TNFSE2 after handling the data imbalance. Table 3 contains the results of MWMOTE balanced dataset and embedding used with various classifiers. Table 4 contains the results of TGAN balanced dataset and embedding used with various classifiers.

**Table 3 - Results of MWMOTE balanced dataset + embedding with various classifiers**

| Embedding | Classifier | Model | AUC-ROC | Sensitivity | Specificity | Precision | False Positive Rate | False Discovery Rate | Accuracy | F1 Score |
|---|---|---|---|---|---|---|---|---|---|---|
| **Isomap** | **DT** | **M16** | 0.9649 | **0.9995** | 0.9303 | 0.9356 | 0.0697 | 0.0644 | 0.9651 | 0.9665 |
| | **RF** | **M17** | 0.9752 | 0.9986 | 0.9517 | 0.9544 | 0.0483 | 0.0456 | 0.9753 | 0.9760 |
| | **XGBoost** | **M18** | 0.7942 | 0.9199 | 0.6685 | 0.7375 | 0.3315 | 0.2625 | 0.7950 | 0.8186 |
| | **LightGBM** | **M19** | 0.8639 | 0.9462 | 0.7815 | 0.8143 | 0.2185 | 0.1857 | 0.8644 | 0.8753 |
| | **GBM** | **M20** | 0.7947 | 0.9043 | 0.6852 | 0.7441 | 0.3148 | 0.2559 | 0.7954 | 0.8164 |
| **MDS** | **DT** | **M21** | 0.5917 | 0.5839 | 0.5994 | 0.5960 | 0.4006 | 0.4040 | 0.5916 | 0.5899 |
| | **RF** | **M22** | 0.6298 | 0.5878 | 0.6718 | 0.6445 | 0.3282 | 0.3555 | 0.6296 | 0.6148 |
| | **XGBoost** | **M23** | 0.6778 | 0.6835 | 0.6721 | 0.6785 | 0.3279 | 0.3215 | 0.6778 | 0.6810 |
| | **LightGBM** | **M24** | 0.6754 | 0.6709 | 0.6799 | 0.6796 | 0.3201 | 0.3204 | 0.6754 | 0.6752 |
| | **GBM** | **M25** | 0.6770 | 0.6769 | 0.6771 | 0.6797 | 0.3229 | 0.3203 | 0.6770 | 0.6783 |
| **t-SNE** | **DT** | **M26** | 0.9623 | 0.9981 | 0.9264 | 0.9321 | 0.0736 | 0.0679 | 0.9625 | 0.9640 |
| | **RF** | **M27** | 0.9706 | 0.9975 | 0.9436 | 0.9471 | 0.0564 | 0.0529 | 0.9708 | 0.9717 |
| | **XGBoost** | **M28** | 0.7661 | 0.8623 | 0.6699 | 0.7256 | 0.3301 | 0.2744 | 0.7667 | 0.7881 |
| | **LightGBM** | **M29** | 0.8412 | 0.9402 | 0.7424 | 0.7870 | 0.2576 | 0.2130 | 0.8419 | 0.8568 |
| | **GBM** | **M30** | 0.7702 | 0.8686 | 0.6718 | 0.7282 | 0.3282 | 0.2718 | 0.7708 | 0.7922 |
| **TNFSE1** | **DT** | **M31** | 0.9885 | 0.9981 | 0.9789 | 0.9795 | 0.0211 | 0.0205 | 0.9885 | 0.9887 |
| | **RF** | **M32** | 0.9915 | 0.9975 | 0.9856 | 0.9859 | 0.0144 | 0.0141 | 0.9916 | 0.9917 |
| | **XGBoost** | **M33** | 0.9684 | 0.9926 | 0.9442 | 0.9474 | 0.0558 | 0.0526 | 0.9685 | 0.9695 |
| | **LightGBM** | **M34** | 0.9883 | 0.9986 | 0.9773 | 0.9788 | 0.0227 | 0.0212 | 0.9882 | 0.9886 |
| | **GBM** | **M35** | 0.9677 | 0.9929 | 0.9425 | 0.9459 | 0.0575 | 0.0541 | 0.9679 | 0.9688 |
| **TNFSE2** | **DT** | **M36** | 0.9883 | 0.9986 | 0.9781 | 0.9788 | 0.0219 | 0.0212 | 0.9884 | 0.9886 |
| | **RF** | **M37** | **0.9917** | 0.9975 | **0.9858** | **0.9862** | **0.0142** | **0.0138** | **0.9917** | **0.9918** |
| | **XGBoost** | **M38** | 0.9707 | 0.9967 | 0.9448 | 0.9481 | 0.0552 | 0.0519 | 0.9709 | 0.9718 |
| | **LightGBM** | **M39** | 0.9872 | 0.9967 | 0.9768 | 0.9782 | 0.0232 | 0.0218 | 0.9870 | 0.9874 |
| | **GBM** | **M40** | 0.9709 | 0.9951 | 0.9467 | 0.9497 | 0.0533 | 0.0503 | 0.9710 | 0.9719 |

**Table 4 - Results of TGAN balanced dataset + embedding with various classifiers**

| Embedding | Classifier | Model | AUC-ROC | Sensitivity | Specificity | Precision | False Positive Rate | False Discovery Rate | Accuracy | F1 Score |
|---|---|---|---|---|---|---|---|---|---|---|
| **Isomap** | **DT** | **M41** | 0.6546 | 0.7646 | 0.5446 | 0.6157 | 0.4554 | 0.3843 | 0.6520 | 0.6821 |
| | **RF** | **M42** | 0.6622 | 0.7470 | 0.5774 | 0.6278 | 0.4226 | 0.3722 | 0.6602 | 0.6822 |
| | **XGBoost** | **M43** | 0.6315 | 0.8212 | 0.4418 | 0.5840 | 0.5582 | 0.4160 | 0.6271 | 0.6826 |
| | **LightGBM** | **M44** | 0.6643 | 0.7317 | 0.5970 | 0.6340 | 0.4030 | 0.3660 | 0.6627 | 0.6793 |
| | **GBM** | **M45** | 0.6338 | 0.8124 | 0.4554 | 0.5873 | 0.5446 | 0.4127 | 0.6297 | 0.6818 |
| **MDS** | **DT** | **M46** | 0.5775 | 0.5782 | 0.5769 | 0.5659 | 0.4231 | 0.4341 | 0.5775 | 0.5720 |
| | **RF** | **M47** | 0.6155 | 0.5551 | 0.6759 | 0.6204 | 0.3241 | 0.3796 | 0.6169 | 0.5860 |
| | **XGBoost** | **M48** | 0.6580 | 0.6671 | 0.6488 | 0.6444 | 0.3512 | 0.3556 | 0.6577 | 0.6556 |
| | **LightGBM** | **M49** | 0.6537 | 0.6782 | 0.6292 | 0.6358 | 0.3708 | 0.3642 | 0.6532 | 0.6563 |

| | | | | | | | | | |
|---|---|---|---|---|---|---|---|---|---|
| | **GBM** | **M50** | 0.6583 | 0.6552 | 0.6615 | 0.6487 | 0.3385 | 0.3513 | 0.6584 | 0.6520 |
| **t-SNE** | **DT** | **M51** | 0.7814 | 0.7800 | 0.7828 | 0.7740 | 0.2172 | 0.2260 | 0.7814 | 0.7770 |
| | **RF** | **M52** | 0.7912 | 0.7789 | 0.8036 | 0.7910 | 0.1964 | 0.2090 | 0.7915 | 0.7849 |
| | **XGBoost** | **M53** | 0.7589 | 0.8076 | 0.7103 | 0.7268 | 0.2897 | 0.2732 | 0.7578 | 0.7650 |
| | **LightGBM** | **M54** | 0.7984 | 0.8957 | 0.7011 | 0.7409 | 0.2989 | 0.2591 | 0.7961 | 0.8110 |
| | **GBM** | **M55** | 0.7619 | 0.8076 | 0.7163 | 0.7309 | 0.2837 | 0.2691 | 0.7609 | 0.7673 |
| **TNFSE1** | **DT** | **M56** | 0.7732 | 0.7683 | 0.7781 | 0.7677 | 0.2219 | 0.2323 | 0.7734 | 0.7680 |
| | **RF** | **M57** | 0.7965 | 0.7851 | 0.8080 | 0.7960 | 0.1920 | 0.2040 | 0.7968 | 0.7905 |
| | **XGBoost** | **M58** | 0.7923 | 0.8891 | 0.6954 | 0.7358 | 0.3046 | 0.2642 | 0.7900 | 0.8053 |
| | **LightGBM** | **M59** | 0.8117 | 0.8775 | 0.7459 | 0.7671 | 0.2541 | 0.2329 | 0.8101 | 0.8186 |
| | **GBM** | **M60** | 0.7983 | 0.8778 | 0.7190 | 0.7488 | 0.2810 | 0.2512 | 0.7965 | 0.8082 |
| **TNFSE2** | **DT** | **M61** | 0.8488 | 0.8621 | 0.8356 | 0.8335 | 0.1644 | 0.1665 | 0.8486 | 0.8476 |
| | **RF** | **M62** | 0.8441 | 0.8857 | 0.8025 | 0.8106 | 0.1975 | 0.1894 | 0.8432 | 0.8465 |
| | **XGBoost** | **M63** | 0.8694 | 0.9659 | 0.7730 | 0.8024 | 0.2270 | 0.1976 | 0.8672 | 0.8766 |
| | **LightGBM** | **M64** | 0.8737 | 0.9227 | 0.8248 | 0.8340 | 0.1752 | 0.1660 | 0.8726 | 0.8761 |
| | **GBM** | **M65** | 0.8219 | 0.9434 | 0.7003 | 0.7502 | 0.2997 | 0.2498 | 0.8190 | 0.8358 |

From Table 3, we can see that M37 which is a combination of MWMOTE, TNFSE2 and Random Forest has the best values of Accuracy (0.9917), F1-Score (0.9918) and AUC-ROC (0.9917) metrics compared to any other models.

The AUC and F1-score were obtained by considering the range of dimensions of embedding from 1 to 20 which is shown in Figure 2. It may be observed from the graph that there is no significant improvement in AUC or F1-score after 16 dimensions. Thus, for the TNFSE2 embedding, 16 dimensions were used.

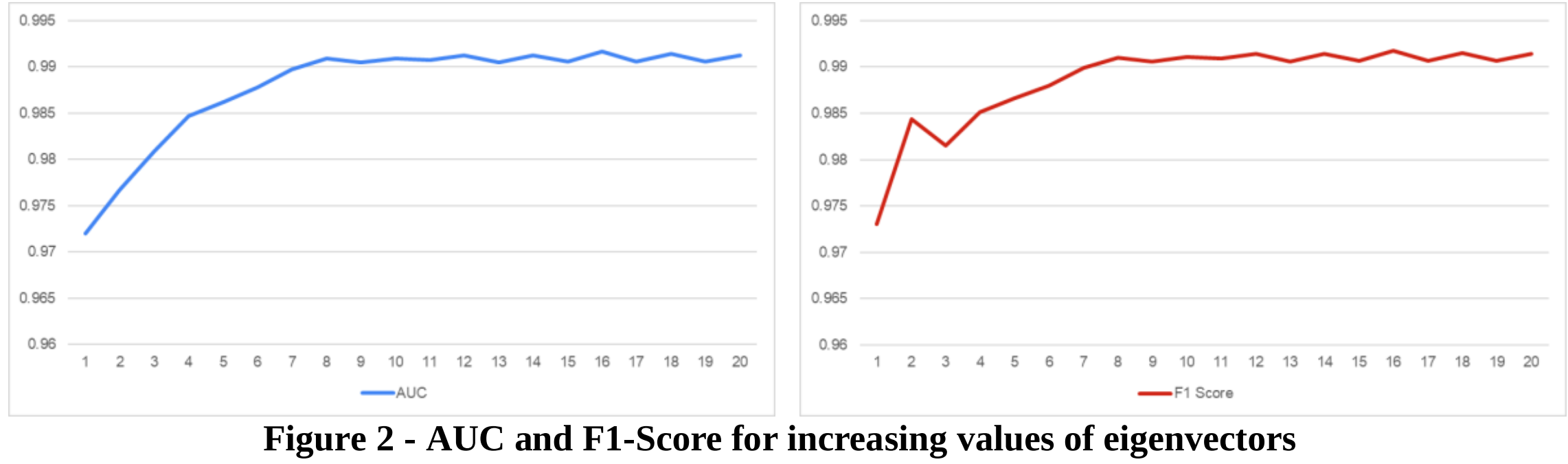

**Figure 2 - AUC and F1-Score for increasing values of eigenvectors**

## 5.4 Business interpretation explaining the effectiveness of TNFSE2 embedding

The results showed that the model performed better when TNFSE2 embedding was done on the dataset. This can be attributed to multiple factors that come into play during the process of embedding. For instance, when the vicinity graph is created for the dataset, edges are formed between the data points whose Euclidean distance is lesser than a fixed $\varepsilon$. This indicates that there is a similarity between some pairs of claims with minor differences in attributes.

In the TNFSE2 model, the affinity matrix is constructed by considering three topological node features – degree, clustering coefficient and summation index. Figure 3 depicts visualization of the vicinity graph for a claim C0.

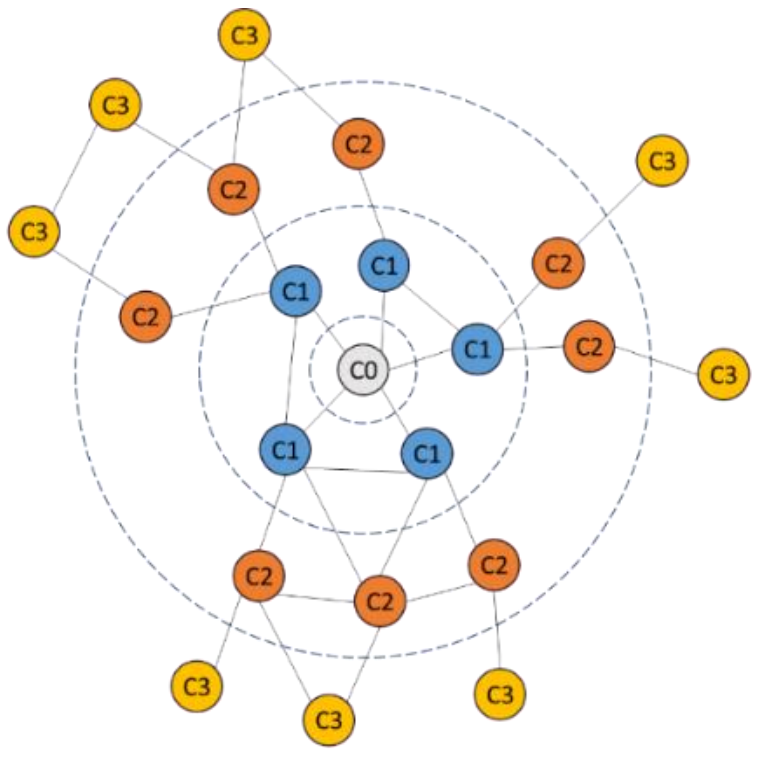

**Figure 3 - Visualization of the vicinity graph for a claim C0**

- Degree of a node C0 represents the number of claims with similar characteristics to C0, here the degree would be the count of the number of C1,
- Clustering Coefficient ($\phi_p$) for claim C0 is the number of edges among C1, in business terms, it means similar claims characteristics like age, date of submission of claims, claim amount, etc. and
- Summation Index ($SI$) is

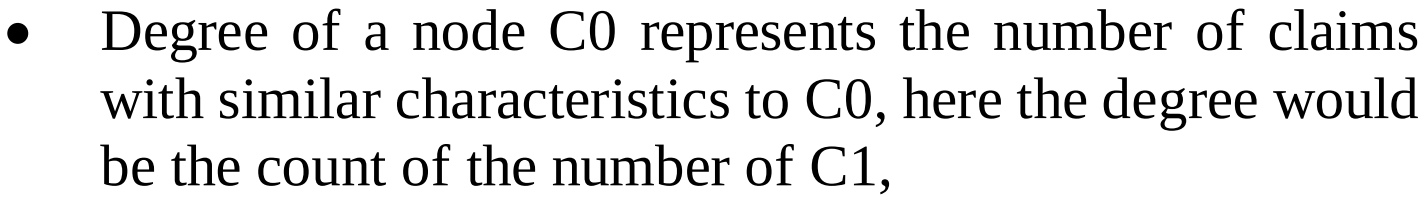

  $SI_i(C0) = degree(C0) \quad if \; i = 0, \quad \sum_{q \in N(C0)} SI_{i-1}(q) \quad if \;\; i > 0.$

  For every claim, we calculate $(SI_0, SI_1, SI_2)$.

Thus, claims with similar characteristics are taken in three levels C0, C1 and C2. The inputs from these topological nodes help in calculating the affinity matrix. All these nodes are defined such that the higher value of $\alpha_{ij}$ indicates a higher correlation (affinity) between two claim records i and j and vice versa. As this embedding model provides multi-level ability to classify claims with similar characteristics, it performed better in classifying fraud and non-fraud cases in comparison to other methods of embedding.

## 6. Conclusion and Future Work

In this paper, we have proposed a three-phased method for building an effective fraud detection model for automobile insurance. Various combination of models in each of the three phases were used to build a fraud detection model and were tested on automobile insurance dataset. The results showed that the combination of MWMOTE, TNFSE2 and Random Forest in phases I, II and III respectively gave the best performance in terms of AUC, Accuracy and F1-Score. As part of future work, the methods can be used for other lines of business like health insurance and an appropriate combination of models in phase I, II & III can be identified for the most effective fraud detection model.

## Conflict of Interest

The authors declare that there is no conflict of interest in this publication

## Acknowledgement

The authors would like to express their thanks to Bhagawan Sri Sathya Sai Baba, founder chancellor, Sri Sathya Sai Institute of Higher Learning.

## References

Arias-Castro, E., Lerman, G., & Zhang, T. (2017). Spectral clustering based on local PCA. *Journal of Machine Learning Research*.

Ashrapov, I. *Tabular GANs for uneven distribution*. , (2020).

Barua, S., Islam, M. M., Yao, X., & Murase, K. (2014). MWMOTE--Majority Weighted Minority Oversampling Technique for Imbalanced Data Set Learning. *IEEE Transactions on Knowledge and Data Engineering*, *26*(2), 405–425. https://doi.org/10.1109/TKDE.2012.232

Belkin, M., & Niyogi, P. (2002). Laplacian Eigenmaps and Spectral Techniques for Embedding and Clustering. In *Advances in Neural Information Processing Systems 14*. https://doi.org/10.7551/mitpress/1120.003.0080

Bermúdez, L., Pérez, J. M., Ayuso, M., Gómez, E., & Vázquez, F. J. (2008). A Bayesian dichotomous model with asymmetric link for fraud in insurance. *Insurance: Mathematics and Economics*, *42*(2), 779–786. https://doi.org/10.1016/j.insmatheco.2007.08.002

Chintalapati, L. S., & Rachakonda, R. S. (2019). Enhanced Affinity for Spectral Clustering using Topological Node Features (TNFS). *International Journal of Engineering and Advanced Technology*, *9*(1), 974–987. https://doi.org/10.35940/ijeat.A9450.109119

Dahm, N., Bunke, H., Caelli, T., & Gao, Y. (2015). Efficient subgraph matching using topological node feature constraints. *Pattern Recognition*, *48*(2), 317–330. https://doi.org/10.1016/j.patcog.2014.05.018

Guelman, L. (2012). Gradient boosting trees for auto insurance loss cost modeling and prediction. *Expert Systems with Applications*, *39*(3), 3659–3667. https://doi.org/10.1016/j.eswa.2011.09.058

Gupta, R. Y., Mudigonda, S. S., & Baruah, P. K. (2021). *TGANs with Machine Learning Models in Automobile Insurance Fraud Detection and Comparative Study with Other Data Imbalance Techniques*. (5), 236–244. https://doi.org/10.35940/ijrte.E5277.019521

Gupta, R. Y., Sai Mudigonda, S., Kandala, P. K., & Baruah, P. K. (2019). Implementation of a Predictive Model for Fraud Detection in Motor Insurance using Gradient Boosting Method and Validation with Actuarial Models. *2019 IEEE International Conference on Clean Energy and Energy Efficient Electronics Circuit for Sustainable Development (INCCES)*, 1–6. https://doi.org/10.1109/INCCES47820.2019.9167733

Insurance Fraud — FBI. (n.d.). Retrieved January 29, 2020, from https://www.fbi.gov/stats-services/publications/insurance-fraud

Japkowicz, N., & Stephen, S. (2002). The class imbalance problem: A systematic study1. *Intelligent Data Analysis*, *6*(5), 429–449. https://doi.org/10.3233/IDA-2002-6504

Kruskal, J. B. (1964a). Multidimensional scaling by optimizing goodness of fit to a nonmetric hypothesis. *Psychometrika*. https://doi.org/10.1007/BF02289565

Kruskal, J. B. (1964b). Nonmetric multidimensional scaling: A numerical method. *Psychometrika*. https://doi.org/10.1007/BF02289694

Kruskal, J., & Wish, M. (2011). Multidimensional Scaling. In *Multidimensional Scaling*. https://doi.org/10.4135/9781412985130

Nian, K., Zhang, H., Tayal, A., Coleman, T., & Li, Y. (2016). Auto insurance fraud detection using unsupervised spectral ranking for anomaly. *The Journal of Finance and Data Science*, *2*(1), 58–75. https://doi.org/10.1016/j.jfds.2016.03.001

O’Connell, A. A., Borg, I., & Groenen, P. (1999). Modern Multidimensional Scaling: Theory and Applications. *Journal of the American Statistical Association*. https://doi.org/10.2307/2669710

Phua, C., Alahakoon, D., & Lee, V. (2004). Minority report in fraud detection. *ACM SIGKDD Explorations Newsletter*, *6*(1), 50–59. https://doi.org/10.1145/1007730.1007738

Powers, D. M. W. (2007). Evaluation: from precision, recall and f-factor. *Technical Report SEI-*

*07-001*.

Rai, N., Baruah, P. K., Mudigonda, S. S., & Kandala, P. K. (2018). Fraud Detection Supervised Machine Learning Models for an Automobile Insurance. *International Journal of Scientific and Engineering Research (IJSER)*, *9*(11), 473–479.

Šubelj, L., Furlan, Š., & Bajec, M. (2011). An expert system for detecting automobile insurance fraud using social network analysis. *Expert Systems with Applications*, *38*(1), 1039–1052. https://doi.org/10.1016/j.eswa.2010.07.143

Subudhi, S., & Panigrahi, S. (2020). Use of optimized Fuzzy C-Means clustering and supervised classifiers for automobile insurance fraud detection. *Journal of King Saud University - Computer and Information Sciences*, *32*(5), 568–575. https://doi.org/10.1016/j.jksuci.2017.09.010

Sundarkumar, G. G., Ravi, V., & Siddeshwar, V. (2015). One-class support vector machine based undersampling: Application to churn prediction and insurance fraud detection. *2015 IEEE International Conference on Computational Intelligence and Computing Research (ICCIC)*, (ii), 1–7. https://doi.org/10.1109/ICCIC.2015.7435726

Tenenbaum, J. B. (2000). A Global Geometric Framework for Nonlinear Dimensionality Reduction. *Science*, *290*(5500), 2319–2323. https://doi.org/10.1126/science.290.5500.2319

Tin Kam Ho. (1995). Random decision forests. *Proceedings of 3rd International Conference on Document Analysis and Recognition*, *1*, 278–282. https://doi.org/10.1109/ICDAR.1995.598994

Van Der Maaten, L., & Hinton, G. (2008). Visualizing data using t-SNE. *Journal of Machine Learning Research*.

VIAENE, S., DEDENE, G., & DERRIG, R. (2005). Auto claim fraud detection using Bayesian learning neural networks. *Expert Systems with Applications*, *29*(3), 653–666. https://doi.org/10.1016/j.eswa.2005.04.030

Wang, J., & Wang, J. (2012). Isomaps. In *Geometric Structure of High-Dimensional Data and Dimensionality Reduction*. https://doi.org/10.1007/978-3-642-27497-8_8

Xu, L., & Veeramachaneni, K. (2018). Synthesizing Tabular Data using Generative Adversarial Networks. *ArXiv*. Retrieved from http://arxiv.org/abs/1811.11264

Xu, W., Wang, S., Zhang, D., & Yang, B. (2011). Random Rough Subspace Based Neural Network Ensemble for Insurance Fraud Detection. *2011 Fourth International Joint Conference on Computational Sciences and Optimization*, 1276–1280. https://doi.org/10.1109/CSO.2011.213

Zhang, X., Li, J., & Yu, H. (2011). Local density adaptive similarity measurement for spectral clustering. *Pattern Recognition Letters*, *32*(2), 352–358. https://doi.org/10.1016/j.patrec.2010.09.014

Žižka, J., Dařena, F., & Svoboda, A. (2019). Random Forest. In *Text Mining with Machine Learning* (pp. 193–200). https://doi.org/10.1201/9780429469275-8